\documentclass{article}
\usepackage{spconf}
\usepackage{amsmath,amssymb,graphicx}
\usepackage{enumitem}
\usepackage{float}
\usepackage{xcolor}
\usepackage[normalem]{ulem}
\usepackage{url}
\usepackage{hyperref}
\setlist{nosep, leftmargin=14pt}

\title{Handling Supervision Scarcity in Chest X-Ray Classification: Long-Tailed and Zero-Shot Learning}

\name{%
\begin{tabular}{c}
Ha-Hieu Pham$^{1,2,3}$ \qquad
Hai-Dang Nguyen$^{3,6}$\sthanks{These authors contributed equally.} \qquad
Thanh-Huy Nguyen$^{4\ast}$ \qquad
Min Xu$^{4}$ \\
Ulas Bagci$^{5}$ \qquad
Trung-Nghia Le$^{1,2}$ \qquad
Huy-Hieu Pham$^{3,6}$\sthanks{Corresponding Author: Huy-Hieu Pham (hieu.ph@vinuni.edu.vn).}%
\end{tabular}%
}

\address{
$^{1}$ University of Science, Ho Chi Minh City, Vietnam \\
$^{2}$ Vietnam National University, Ho Chi Minh City, Vietnam \\
$^{3}$ VinUni-Illinois Smart Health Center, VinUniversity, Hanoi, Vietnam \\
$^{4}$ Carnegie Mellon University, Pittsburgh, PA, USA \\
$^{5}$ Northwestern University, Chicago, IL, USA \\
$^{6}$ College of Engineering \& Computer Science, VinUniversity, Hanoi, Vietnam
}

\begin{document}
%
\maketitle

\begin{abstract}
\label{abs}
Chest X-Ray (CXR) classification in clinical practice is often limited by imperfect supervision, arising from (i) extreme long-tailed multi-label disease distributions and (ii) missing annotations for rare or previously unseen findings. The CXR-LT 2026 challenge addresses these issues on a PadChest-based benchmark with a 36-class label space split into 30 in-distribution classes for training and 6 out-of-distribution (OOD) classes for zero-shot evaluation. We present task-specific solutions tailored to the distinct supervision regimes. For Task~1 (long-tailed multi-label classification), we adopt an imbalance-aware multi-label learning strategy to improve recognition of tail classes while maintaining stable performance on frequent findings. For Task~2 (zero-shot OOD recognition), we propose a prediction approach that produces scores for unseen disease categories without using any supervised labels or examples from the OOD classes during training. Evaluated with macro-averaged mean Average Precision (mAP), our method achieves strong performance on both tasks, ranking first on the public leaderboard of the development phase. Code and pre-trained models are available at \url{https://github.com/hieuphamha19/CXR\_LT}.
\end{abstract}


\section{Introduction}
\label{sec:intro}

Chest X-Ray (CXR) imaging is one of the most widely used diagnostic tools in clinical practice, and deep learning has demonstrated strong potential for automated CXR interpretation \cite{10.1371/journal.pmed.1002686}. Despite this progress, reliable deployment remains challenging because supervision in large scale CXR datasets is often imperfect, with noisy or incomplete labels \cite{mimic-cxr}. In particular, disease annotations typically follow an extreme long-tailed distribution, where a small number of common findings dominate the data while many clinically important abnormalities appear infrequently, causing standard learning approaches to favor head classes and underperform on rare tail findings \cite{YANG2026112683}. This issue is further compounded by the multi-label nature of CXR diagnosis, in which multiple abnormalities may co-occur within the same image, increasing the difficulty of learning balanced and well calibrated predictions.

To facilitate progress on these clinically realistic challenges, the CXR-LT challenge series has established benchmarks for long-tailed, multi-label CXR classification and zero-shot generalization. CXR-LT 2023 introduced a long-tailed benchmark based on MIMIC-CXR with 26 disease findings, while CXR-LT 2024 expanded the label space to 45 findings across more than 377{,}000 images and incorporated zero-shot recognition of unseen diseases~\cite{mimic-cxr,cxr_lt_2023,cxr_lt_2024}. Building on this line of work, CXR-LT 2026\footnote{\url{https://cxr-lt.github.io/CXR-LT-2026}} evaluates long-tailed multi-label classification and zero-shot recognition of unseen findings within the same imaging domain.


In this work, we propose an imbalance-aware multi-label learning strategy that improves recognition of rare (tail) findings while maintaining competitive performance on frequent classes. We additionally develop a zero-shot prediction approach for out-of-distribution (OOD) diseases that produces scores for unseen findings without using any OOD labels during training. The proposed solutions are designed to be robust to label imbalance and incomplete supervision, combining training-time reweighting with lightweight inference-time refinement. We evaluated our method on the CXR-LT 2026 benchmark and achieved strong performance on both tasks, ranking first on the public leaderboard at the end of the development phase under the official evaluation protocol.

\section{CXR-LT 2026 Challenge Overview}

The CXR-LT 2026 benchmark is built on a curated subset of the PadChest dataset and enriched with PadChest-GR annotations, resulting in over 160{,}000 chest X-Ray images with manually verified disease labels~\cite{padchest,padchest_gr}. Unlike previous CXR-LT editions derived from MIMIC-CXR, the use of PadChest brings additional diversity in patient population, acquisition protocols, and annotation practices, encouraging methods that generalize beyond a single data source and remain reliable under distribution shifts. The label space includes 36 disease findings, partitioned into 30 in-distribution (ID) classes and 6 out-of-distribution (OOD) classes. The challenge comprises two independent tasks that reflect complementary real-world needs:
\begin{itemize}
    \item \textbf{Task 1 (Long-tailed multi-label classification).} Models are trained and evaluated on the 30 ID findings, but must remain robust under severe class imbalance, strong label co-occurrence, and heterogeneous acquisition settings.
    \item \textbf{Task 2 (Zero-shot OOD recognition).} Models must produce predictions for the 6 OOD findings without using any supervised labels or training examples from these categories during training, highlighting the ability to transfer knowledge to unseen clinical concepts.
\end{itemize}

Participants are provided with a large automatically labeled training set, along with unlabeled validation and test sets. Performance is assessed using macro-averaged mean Average Precision (mAP), which weights each class equally and therefore emphasizes balanced performance across both frequent and rare findings rather than improvements dominated by head classes.

\section{Methodology}

\subsection{Task 1: Long-tailed multi-label classification}
\label{subsec:task1}

Task~1 requires multi-label classification over 30 in-distribution findings under a long-tailed label distribution. Our approach combines imbalance-aware training for robust tail recognition and an inference-time pipeline consisting of TTA, weighted ensembling, and a post-processing step.

\noindent\textbf{Preprocessing.}
We load radiographs with OpenCV. We apply percentile-based intensity clipping followed by rescaling to obtain a consistent dynamic range. Each image is resized to $512\times512$, converted to 3-channel RGB (by channel replication for grayscale images), and normalized with ImageNet mean and standard deviation before being fed to the network.

\noindent\textbf{Model architecture.}
We use ConvNeXtV2-Base as the backbone, initialized from weights pre-trained on chest X-Ray images from MIMIC-CXR to provide representations. We then fine-tune two complementary variants that differ only in the classification head: (i) a standard MLP head and (ii) a CSRA head that incorporates class-specific spatial attention \cite{woo2023convnext,Zhu_2021_ICCV}. Both models output $C{=}30$ logits $z\in\mathbb{R}^{C}$ and apply sigmoid activation to produce multi-label probabilities:

\[
p_c = \sigma(z_c), \quad c=1,\dots,C.
\]

\noindent\textbf{Imbalance-aware objective (Distribution-Balanced loss).}
To mitigate extreme class imbalance, we adopt a Distribution-Balanced (DB) loss that combines (i) class-dependent re-weighting based on effective-number statistics and (ii) a margin adjustment for positive labels. Given label $y_c\in\{0,1\}$ and logit $z_c$, we first apply a positive margin
\[
z'_c = z_c - y_c\, m_c,
\]
then optimize a reweighted BCE objective
\[
\mathcal{L}_{\text{DB}} = \frac{1}{C}\sum_{c=1}^{C} w_c \cdot \mathrm{BCEWithLogits}(z'_c, y_c).
\]
The per-class weight $w_c$ is derived from the effective number of samples using class counts $n_c$
\[
\mathrm{eff}_c = \frac{1-\beta}{1-\beta^{n_c}},\qquad
w_c \propto \left(\mathrm{eff}_c\right)^{\alpha},
\]
where $\beta$ is fixed to $0.9999$ and $\alpha$ controls the rebalancing strength. This design increases the contribution of tail classes while preventing unstable over-amplification. 

\noindent\textbf{Class-Aware Sampling (CAS).}
In addition to loss reweighting, we use a repeat-factor style sampler to oversample images containing rare positive labels. Let $f_c$ denote the empirical frequency of class $c$ and $T$ a frequency threshold. We compute a class-wise repeat factor
\[
r(c)=\max\!\left(1,\sqrt{\frac{T}{f_c}}\right),
\]
and assign each sample $i$ a repeat factor based on the rarest positive class it contains
\[
r_i=\min\!\left(r_{\max},\, \max_{c:\, y_{i,c}=1} r(c)\right),
\]
with an upper cap $r_{\max}$ for stability. This sampling strategy increases exposure to tail positives without excessively distorting the data distribution.

\noindent\textbf{Ensemble inference with TTA.}
For the final submission, we ensemble two trained checkpoints using a weighted mean (raw weights $a_1{=}1.0$ and $a_2{=}1.5$, normalized internally). At inference time, we apply test-time augmentation (TTA) by averaging predictions over a small set of label-preserving transforms: identity, horizontal flip, rotations by $\pm 5^\circ$, and mild zoom-in/zoom-out (scaling factors $1.1$ and $0.9$ with center crop/padding to keep the original resolution). For model $k$ and augmentation set $\mathcal{T}$, the TTA-merged prediction is

\[
p^{(k)}_{\mathrm{tta}}(x)=\frac{1}{|\mathcal{T}|}\sum_{t\in\mathcal{T}} \sigma\!\left(z^{(k)}(t(x))\right),
\]
where $z^{(k)}(\cdot)$ denotes the model logits and $\sigma(\cdot)$ is the element-wise sigmoid. The final ensemble prediction is
\[
p_{\mathrm{ens}}(x)=\sum_{k}\tilde{a}_k\, p^{(k)}_{\mathrm{tta}}(x),\qquad
\tilde{a}_k=\frac{a_k}{\sum_j a_j}.
\]

\noindent\textbf{Post-processing via normal gating.}
We apply a lightweight post-hoc refinement using the predicted probability of the \textit{Normal} class $p_0$. For each abnormal class $c\neq 0$, we suppress its score by
\[
p_c \leftarrow p_c \cdot (1-p_0)^{\alpha_{\text{ng}}},
\]
where $\alpha_{\text{ng}}=0.5$ in our submission. Intuitively, when the model is highly confident that an image is normal, abnormal probabilities should be jointly down-weighted. This operation reduces spurious positives while preserving the relative ranking among abnormal findings.

\subsection{Task 2: Zero-shot OOD detection}
\label{subsec:task2}

Task~2 evaluates zero-shot recognition of six out-of-distribution (OOD) findings:
\textit{Scoliosis, Osteopenia, Bulla, Infarction, Adenopathy,} and \textit{Goiter}.
Because no supervised examples are provided for these categories, we formulate Task~2
as a vision-language matching problem and infer OOD probabilities via image-text similarity.

\noindent\textbf{Model.}
We adopt \textbf{WhyXrayCLIP}, a CXR-specialized vision-language model built upon an OpenCLIP ViT-L/14 backbone and further fine-tuned on large-scale chest X-Ray image-report pairs from MIMIC-CXR \cite{NEURIPS2024_a4e683f0}. Through contrastive image-text learning on radiology reports, WhyXrayCLIP learns CXR-specific visual semantics and their language correspondences (e.g., abnormalities, anatomical regions, and descriptive patterns), enabling more accurate alignment between radiographs and textual descriptions. This domain-adapted alignment makes classification naturally prompt-driven: an image can be scored directly against textual descriptions of candidate findings without requiring task-specific supervision. As a result, the model is particularly suitable for Task~2, where training on OOD labels is prohibited, as it supports robust zero-shot recognition by leveraging text prompts as class prototypes.

\noindent\textbf{Preprocessing.}
Each CXR is loaded in grayscale with OpenCV, cast to \texttt{float32}, and normalized to
$[0,1]$ while handling both 8-bit and 16-bit images (division by $255$ or $65535$
depending on the dynamic range). Since CLIP expects three channels, we replicate the
grayscale image to form a 3-channel input, resize to $224\times224$, and apply CLIP-style
mean/std normalization.

\noindent\textbf{Prompt ensembling.}
For each OOD category, we define a small set of generic radiological text descriptions
${t_{c,k}}_{k=1}^{K}$. These prompts are fixed and used only at inference time. Each
prompt is encoded into a unit-normalized text embedding, and prompt ensembling is employed
to improve robustness to wording variations. The full prompt list is provided in the code (link in the Abstract).

\noindent\textbf{Zero-shot scoring and probability mapping.}
For an input radiograph, WhyXrayCLIP produces a unit-normalized image embedding
$\hat{v}$. For each class $c$, we compute cosine similarities between $\hat{v}$ and
its $K$ prompt embeddings $\{\hat{t}_{c,k}\}_{k=1}^{K}$, average them across prompts,
and map the resulting score to $[0,1]$ using a scaled sigmoid
\[
p_c = \sigma\!\left(\alpha \cdot \frac{1}{K}\sum_{k=1}^{K}\left\langle \hat{v},\, \hat{t}_{c,k}\right\rangle\right),
\]
where $\sigma(\cdot)$ denotes the sigmoid function and $\alpha$ is a fixed scaling
factor (e.g., $\alpha=5$) used to sharpen the separation between low- and high-similarity
cases. This procedure yields six OOD probabilities per image.

\noindent\textbf{Inference and submission.}
We pre-compute and cache prompt embeddings once, then perform batch-wise GPU inference to
compute image embeddings and class scores efficiently via matrix multiplication. Final
predictions are reported and submitted as probabilities for the six OOD findings in the
predefined order.

\section{Results}
\label{sec:results}

We report results on the public development phase of the CXR-LT 2026 challenge. The official primary metric is macro-averaged mAP, which equally weights all labels and thus better reflects performance under long-tailed multi-label distributions, especially for rare findings. We additionally report mAUC, mF1, and mECE to characterize discrimination, thresholded decision quality, and calibration, providing a more complete view beyond the primary metric. 

\begin{table}[ht]
\centering
\scriptsize
\setlength{\tabcolsep}{6pt}
\begin{tabular}{r l c c c c}
\hline
\# & Participant & mAP $\uparrow$ & mAUC $\uparrow$ & mF1 $\uparrow$ & mECE $\downarrow$ \\
\hline
1  & CVMAIL x MIHL & \textbf{0.583} & \uline{0.919} & \textbf{0.376} & 0.928 \\
2  & uccaeid                 & \uline{0.535} & 0.895 & 0.203 & \textbf{0.620} \\
3  & mshamani                & 0.531 & 0.903 & 0.196 & 0.637 \\
4  & laghaei                 & 0.530 & 0.902 & 0.196 & 0.640 \\
5  & ahajighasem             & 0.530 & 0.891 & 0.204 & 0.632 \\
6  & MLtests                 & 0.524 & 0.895 & 0.206 & \uline{0.629} \\
7  & Nikhil Rao Sulake       & 0.522 & 0.896 & 0.266 & 0.891 \\
8  & 5Gmodels                & 0.520 & 0.880 & 0.203 & \textbf{0.620} \\
9  & hannah                  & 0.518 & \textbf{0.920} & \uline{0.369} & 0.927 \\
10 & kzan25                  & 0.515 & 0.908 & 0.203 & 0.641 \\
\hline
\end{tabular}
\caption{Public development leaderboard for CXR-LT 2026 Task~1 (Top-10). Best and second-best results are highlighted in \textbf{bold} and \uline{underlined}.}
\label{tab:lb_task1_nodateid}
\end{table}

\subsection{Task 1: Long-tailed multi-label classification}

As shown in Table~\ref{tab:lb_task1_nodateid}, our submission ranks \textbf{1st} on the development leaderboard with a macro mAP of \textbf{0.583}, outperforming the runner-up (0.535) by \textbf{0.048}. We also achieve a mAUC of 0.919 and the best mF1 of \textbf{0.376}, indicating competitive discrimination and thresholded predictions under long-tailed multi-label evaluation. However, the high mECE (0.928) suggests suboptimal calibration on the public split.

\begin{table}[ht]
\centering
\scriptsize
\setlength{\tabcolsep}{6pt}
\begin{tabular}{r l c c c c}
\hline
\# & Participant & mAP $\uparrow$ & mAUC $\uparrow$ & mF1 $\uparrow$ & mECE $\downarrow$ \\
\hline
1  & CVMAIL x MIHL & \textbf{0.467} & \textbf{0.779} & 0.259 & \uline{0.516} \\
2  & hannah        & \uline{0.365} & 0.629 & \textbf{0.284} & \textbf{0.169} \\
3  & uccaeid       & 0.337 & \uline{0.693} & 0.181 & 0.691 \\
4  & zuang         & 0.337 & 0.587 & 0.259 & 0.784 \\
5  & MLtests       & 0.326 & 0.671 & 0.198 & 0.694 \\
6  & IDEAMCVG      & 0.326 & 0.636 & \uline{0.280} & 0.638 \\
7  & ahajighasem   & 0.309 & 0.647 & 0.181 & 0.689 \\
8  & farhankhan    & 0.291 & 0.636 & 0.262 & 0.655 \\
9  & laghaei       & 0.269 & 0.593 & 0.000 & 0.793 \\
10 & IAU CV team   & 0.260 & 0.543 & 0.066 & 0.792 \\
\hline
\end{tabular}
\caption{Public development leaderboard for CXR-LT 2026 Task~2 (Top-10). Best and second-best results are highlighted in \textbf{bold} and \uline{underlined}.}
\label{tab:lb_task2_nodateid}
\end{table}

\subsection{Task 2: Zero-shot OOD recognition}

For Task~2 (Table~\ref{tab:lb_task2_nodateid}), our method ranks \textbf{1st} with a macro mAP of \textbf{0.467}, surpassing the runner-up (0.365) by \textbf{0.102}, and achieves the best mAUC (\textbf{0.779}). This indicates stronger zero-shot ranking on unseen OOD findings. Although calibration remains imperfect, our mECE (0.516) is the \uline{second lowest} among the Top-10 on the public split. Results are reported on the public development phase; the hidden test set will better reflect generalization.

\section{Conclusion}
\label{sec:conclusion}

We presented task-specific solutions for the CXR-LT 2026 challenge under two supervision regimes: long-tailed multi-label classification of in-distribution findings (Task~1) and zero-shot recognition of out-of-distribution (OOD) findings (Task~2). Our imbalance-aware multi-label learning improves performance on rare diseases while preserving accuracy on frequent findings, and our zero-shot OOD inference enables prediction of unseen disease categories without using any OOD labels or examples during training. On the public development leaderboard, our method ranks 1st on both tasks, achieving 0.583 mAP on Task~1 and 0.467 mAP on Task~2.

\noindent\textbf{Limitations.} Our evaluation is limited to the public development set, since the official test set was not released at the time of submission; thus, generalization on the hidden test benchmark remains to be verified.

\noindent\textbf{Future work.} We will improve calibration and robustness across sites and acquisition settings, investigate stronger text-image alignment and label semantics for better zero-shot transfer, and extend the framework to radiology vision-language tasks such as report generation and medical visual question answering (VQA) \cite{nguyen2025vindrcxrvqavisualquestionanswering}.

\section{Acknowledgments}
\label{sec:acknowledgments}

 This work was supported by the National Foundation for Science and Technology Development (NAFOSTED) through Project IZVSZ2-229539 (2025–2027).

\bibliographystyle{IEEEbib}
\bibliography{refs}

\end{document}